\documentclass[letterpaper, 10 pt, conference]{ieeeconf}  % Comment this line out if you need a4paper
\usepackage{cite}
\usepackage[pdftex]{graphicx}
\usepackage{amsmath}
\usepackage{amssymb}  % assumes amsmath package installed

\usepackage{epsfig} % for postscript graphics files
\usepackage{flushend}
\usepackage{url}
\newcommand{\reva}{}
\newcommand{\revb}{}
\newcommand{\revc}{}
\newcommand{\revd}{}
\newcommand{\reve}{}%[1]{\textcolor{red}{#1}}
\newcommand{\figurespace}{}%\vspace{-3mm}}

\IEEEoverridecommandlockouts                              % This command is only needed if 

\title{Warped Hypertime Representations for Long-term Autonomy of Mobile Robots}

\author{Tom{\'a}{\v s} Krajn{\'i}k$^1$, Tom{\'a}{\v s} Vintr$^1$, Sergi Molina$^2$, Jaime Pulido Fentanes$^2$,\\ Grzegorz Cielniak$^2$, Oscar Martinez Mozos$^3$, George Broughton$^1$, Tom Duckett$^2$
	\thanks{The work is funded by the CSF project 17-27006Y STRoLL, EU H2020 project 732737 ILIAD, by the Spanish program RYC-2014-15029 and OP VVV MEYS RCI project CZ.02.1.01/0.0/0.0/16\_019/0000765.}%
	\thanks{$^1$ Artificial Intelligence Center, Faculty of Electrical Engineering, Czech Technical University, CZ {\tt\footnotesize{name.surname@fel.cvut.cz}}}%
	\thanks{$^2$ Lincoln Centre for Autonomous Systems, University of Lincoln, UK}%
	\thanks{$^3$ Technical University of Cartagena, Spain}%
}
\begin{document}
\maketitle

\begin{abstract}
   This paper presents a novel method for introducing time into discrete and continuous spatial representations used in mobile robotics, by modelling long-term, pseudo-periodic variations caused by human activities or natural processes.
Unlike previous approaches, the proposed method does not treat time and space separately, and its continuous nature respects both the temporal and spatial continuity of the modeled phenomena.
The key idea is to extend the spatial model with a set of wrapped time dimensions that represent the periodicities of the observed events.
By performing clustering over this extended representation, we obtain a model that allows the prediction of probabilistic distributions of future states and events in both discrete and continuous spatial representations.
We apply the proposed algorithm to several long-term datasets acquired by mobile robots and show that the method enables a robot to predict future states of representations with different dimensions.
The experiments further show that the method achieves more accurate predictions than the previous state of the art.
		
\end{abstract}

%\begin{IEEEkeywords}
%Mapping, Learning and Adaptive Systems, Service Robots 
%\end{IEEEkeywords}

   \section{Introduction}
Advances in autonomous robotics are gradually enabling deployment of robots in human-populated environments~\cite{hawes2016strands}.
Human activity tends to cause changes to the environments it takes place in, and the mobile robots that share these environments need to be able to cope with such never-ending changes.
Many authors have shown that environment models which adapt to changes improve the overall ability of mobile robots to operate over longer time periods~\cite{Biber09,markov,kucner2013conditional,fremen,churchill}. 
Since long-term autonomous operation improves the chances of observing the environment changes, mobile robots gain the opportunity to learn the environment structure, but also how it changes over time~\cite{hawes2016strands}.
\begin{figure}[!t]
   \begin{center}
      \includegraphics[width=0.99\columnwidth]{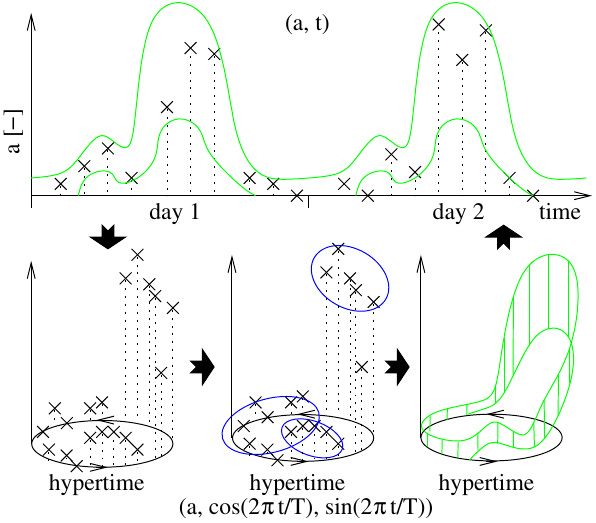}
	   \caption{Method overview: The data points (a,t) observed over time (top, black) are first processed by frequency analysis~\cite{fremen} to determine a dominant periodicity $T$. Then, the time $t$ is projected onto a 2d space (called hypertime) and the vectors $(a,t)$ become $(a,cos(2\pi\,t/T),sin(2\pi\,t/T))$ (bottom, left). The projected data are then clustered (bottom, center, blue) to estimate the distribution of $a$ over the hypertime space (green). Projection of the distribution back to the uni-dimensional time domain allows to calculate the probabilistic distribution of $a$ for any past or future time.}\label{fig:hypertime}
\end{center}
\vspace{-5mm}
\end{figure}

Methods that explicitly represent evolution of the environment over time endow the robots with the ability to predict the future state of the world, which has a positive impact on the efficiency of their long-term operation.
For example, models of movement patterns of objects~\cite{nils} or people~\cite{kucner2013conditional,sergi} improve motion planning, and explicit models of the persistence~\cite{markov,rosen} or periodicity~\cite{fremen} of the environment states improve visual localisation. 
However, the latter methods are applicable to a wider range of scenarios. 
In particular, Frequency Map Enhancement (FreMEn)~\cite{fremen} can introduce dynamics into most discrete environmental representations used in robotics, and  was used to improve robotic mapping~\cite{fremen,2}, localisation~\cite{fremen,4,6}, path planning~\cite{7}, robotic search~\cite{fremen_search}, activity recognition~\cite{9}, patrolling~\cite{10,16}, exploration~\cite{11,12}, task scheduling~\cite{13} and human-robot interaction~\cite{14,15}.
\revb{Moreover, FreMEn allows for temporal-context-based novelty and anomaly detection~\cite{fremen,2,17}, and it was shown to outperform other temporal models both in prediction accuracy and computational efficiency~\cite{fremen_search,9,10,12,15,17}.}
FreMEn is based on the assumption that some of the mid- to long-term processes that exhibit themselves through environment changes are periodic, e.g. seasonal foliage changes, day-night illumination or human activities characterised by regular routines.
By representing the \revb{dynamics of the individual environment model components in the frequency domain}~\cite{fremen}, FreMEn can efficiently store past observations, and provide accurate predictions of the future environment states.
However, it \revc{was tailored to models} consisting of components with binary states, e.g. occupancy grids, landmarks or topological maps.

In this paper, we present a method that also aims to explicitly model periodic changes of the environment, but it can be applied to continuous, multi-dimensional representations, and can predict the probability distribution of a given random variable at a particular time and location.
The method first uses FreMEn~\cite{fremen} to identify (temporally) periodic patterns in the data gathered.
Then, it transforms each time periodicity into a pair of dimensions that form a circle in $2d$ space and adds (concatenates) these dimensions to the vectors that represent the spatial aspects of the modelled phenomena.
Finally, a generalised model is buil\reva{t} by applying traditional techniques like clustering or expectation-maximisation over the warped time-space representation.
The resulting multi-modal model represents both the structure of the space and temporal patterns of the changes or events as shown in Fig.~\ref{fig:hypertime}.
In this way, the proposed method can turn a spatial representation into a spatio-temporal one by extending it with several wrapped dimensions representing time, with each pair of temporal dimensions representing a given periodicity observed in the gathered data.
We hypothesise that since the proposed model respects the spatio-temporal continuity of the modelled phenomena, it provides more accurate predictions than models which partition the space into discrete elements, or that models which neglect the temporal aspects.

We provide a description of the proposed method and experimental evidence of its capability to efficiently represent spatio-temporal data and to predict future states of the environment.
Unlike the previous works~\cite{markov,rosen,kucner2013conditional,fremen}, which \revc{aim to} introduce time into models that represent the environment by a discrete set of binary states, such as the visibility of landmarks or cell occupancy in grids, our method is able to work with continuous and higher-dimensional variables, e.g.\ robot velocities, object positions, pedestrian flows, etc.
Moreover, the method explicitly represents and predicts not only the value of a given state, but also its probabilistic distribution at a particular time and location, which can be useful for task scheduling and planning~\cite{mudrova_ecmr15}.

Our experiments, based on real world data gathered over several weeks, confirm that the method achieved more accurate predictions than both static models and models that aim to represent time over a discretised space only.

   \section{Related work}

In mobile robotics, the effects of environment variations were studied mainly from the perspective of localisation and mapping, because neglecting the environment change gradually deteriorates the ability of the robot to determine its position reliably and accurately.
Assuming that the world is non-stationary, the authors of~\cite{Biber09,milford_persistent_2010,konolige_towards_2009,hochdorfer_towards_2009,churchill} proposed approaches that create, refine and update world models in a continuous manner.
Furthermore, Ambrus et al.~\cite{rares} demonstrated that the ability of continuous remapping not only allows to refine models of the static environment structure, but also opens up the possibility to learn object models from the spatial changes observed~\cite{discovery}.

Unlike the aforementioned works, which focused on the spatial structure and appearance aspects of the changes observed, other authors~\cite{Biber09,maja,Dayoub11,markov,fremen} focused on modelling the temporal aspects.
For example, \cite{Biber09} and \cite{maja} represent the environment dynamics by multiple temporal models with different timescales, where the best map for localisation is chosen by its consistency with current readings. 
Dayoub et al.~\cite{Dayoub11} and Rosen et al.~\cite{rosen} used statistical methods to create feature persistence models and reasoned about the stability of the environmental states over time. 
Tipaldi et al.~\cite{markov} proposed to represent the occupancy of cells in a traditional occupancy grid with a Hidden Markov Model. 
Krajnik et al.~\cite{fremen} represent the probability of environment states in the spectral domain, which captures cyclic (daily, weekly, yearly) patterns of environmental changes as well as their persistence.

The aforementioned approaches demonstrated that considering temporal aspects (and especially their persistence and periodicity) in robotic models improves not only mobile robot localisation~\cite{fremen,markov,Biber09}, but also planning~\cite{7,fremen_search} and exploration~\cite{11}.
However, these temporal representations were tailored to model the probability of a single state over time, and thus were applied only to individual components of the discretised models, e.g.\ cells in an occupancy grid~\cite{markov,11}, visibility of landmarks~\cite{fremen}, traversability of edges in topological maps~\cite{7} or human presence in a particular room~\cite{fremen_search}.
Since the spatial interdependence of these components was neglected, the above models were actually considering only temporal and not spatial-temporal relations of the represented environments. 
This results not only in memory inefficiency (because of the necessity to model a high number of discrete states separately) but also in the inability of the representation to estimate environment states at locations where no measurements were taken, e.g.\ if a certain cell in an occupancy grid is occluded, its state is unknown even if the neighbouring cells are occupied, because the cell is part of a wall or ground.

Spatio-temporal relations of discrete environment models were investigated in~\cite{kucner2013conditional,wang2014modeling}.
Kuczner et al.~\cite{kucner2013conditional} proposed to model how the occupancy likelihood of a given cell in a grid is influenced by the neighbouring cells and showed that this representation allows to model object movement directly 
in an occupancy grid.
A similar approach was proposed in \cite{wang2014modeling}, where the direction of traversal over each cell is obtained using an input-output Hidden Markov Model connected to neighboring cells.
However, these models represent only local spatial dependencies and suffer from a major disadvantage of the discretised models -- memory inefficiency. 
Therefore, in their latest work, \cite{kucner2017enabling,ransalu2018} model a given set of spatio-temporal phenomena (the motion of people and wind flow) in a continuous domain, building their model by means of Expectation Maximisation. 
Moreover, \cite{kinodynamic} shows how to use this representation for robot motion planning in crowded environments.

O'Callaghan and Ramos~\cite{o2012gaussian} also argue in favour of continuous models, showing the advantages of Gaussian Mixture-based representations in terms of memory efficiency and utility for mobile robot navigation.
\revb{Authors of \cite{ramos2016hilbert} speed up building and updating of the proposed models by using an elegant combination of kernels and optimization methods.
Moreover, the method is extended to perform short-term predictions of the environment state based on the history of recent observations~\cite{ransalu2016ST}.}
%The speed-up achieved allows to recalculate the model relatively quickly, which keeps the model updated with the changes in the robot's operational environment.

Unlike the work of Ramos et al.~\cite{ramos2016hilbert}, which is aimed primarily at modelling the spatial structure, and~\cite{kucner2017enabling}, which aims to make short-term predictions of the motion of people, our aim is to create universal, spatial-temporal models capable of long-term predictions of various phenomena.
Inspired by the ability of the continuous models~\cite{ramos2016hilbert,kucner2017enabling} to represent spatio-temporal phenomena and the predictive power of spectral representations~\cite{fremen}, we propose a novel method which allows to introduce the notion of time into state-of-the-art spatial models used in mobile robotics.
\revb{Unlike methods, that perform predictions based on recent observations, our method can accurately predict the environment states using observations gathered days to months before the prediction.
As demonstrated in~\cite{17}, the predictive accuracy of FreMEn and hypertime exceeds popular, state-of-the-art methods like facebook's Prophet.
Moreover, the STRANDS project~\cite{hawes2016strands} showed that the predictions provided have a positive impact on the efficiency of mobile robot operation in long-term scenarios.
} 

   \section{\revb{Method}}

\subsection{Motivation}

\revb{Let us consider the scenarios of the STRANDS project~\cite{hawes2016strands}, which utilised the FreMEn~\cite{fremen} method, and created robots that achieved four months of autonomy in human-populated environments.
In the STRANDS deployments, the robots had to detect anomalies, provide an info-terminal service, and guide people to designated locations or to other people.
Anomalies were not defined by hard-coded rules, but simply as events and activities occurring at unusual locations and times.}
Thus, the robots had to create models of the usual events and activities from their past experience and use outlier detection to identify anomalies. 
To provide an efficient info-terminal service, a robot has to position itself close to areas with a high level of pedestrian traffic.
However, to avoid causing nuisance to people, it should arrive at these locations before they become crowded.
Thus, the robot has to anticipate the occurrence of people at a given time and location based on its past experience.
To guide people to the designated locations, the robot needs to be able to plan its path so that it avoids areas which might be blocked or congested.
If the robot is supposed to guide someone to another person, it needs to know where that person usually is.
To be able to perform the aforementioned tasks at all, the robot has to be able to localise itself despite the environment changes and to plan its path so that it reaches the desired locations in an efficient way.
\revb{While predicting closed doors, congestions or localisation failures from recent observations is beneficial, long-term predictions based on past experience allow to construct plans that avoid these situations.}

As already demonstrated in the STRANDS project~\cite{hawes2016strands}, the efficiency of a mobile robot improves if it can predict the future states of the environment relevant to the service provided.
However, the predictive engine of the STRANDS robots was based on the FreMEn method, \revc{which, in its original form, aims to} represent independent binary states only.
From a formal point of view, FreMEn represents a given phenomenon by $n$ independent states $s_i$, and can estimate the conditional Bernoulli distribution $p(s_i=1|t)$ of each state for past or future time $t$. %changed position
%The independence of temporal models does not depict the spatio-temporal phenomena precisely as it leads to the sparse model, when the spatial resolution of the model grows. %new
%Although it can be addressed by using spatial ordering~\cite{Cliff1975Model}, the spatial ordering has to be predefined and cannot be changed during the learning process~\cite{Shi2018Machine}. %new
%From a formal point of view, FreMEn represents a given phenomenon by $n$ independent states $s_i$, and can estimate the conditional Bernoulli distribution $p(s_i=1|t))$ of each state for past or future time $t$.
In contrast, the method presented in this paper aims to estimate and predict $p(a|\mathbf{x},t)$, where $a$ is the predicted variable (e.g. landmark visibility, robot velocity, number of people present etc.), and $\mathbf{x},t$ are the location and time.
This not only alleviates the necessity to spatially discretise the modelled phenomenon, but also allows to predict other-than-Bernoulli distributions.

\subsection{Core Idea}

The problem of finding the conditional distribution $p(a|\mathbf{x},t)$ is that while the modelled space $\mathbf{x}\in\mathbb{R}^m$ is constrained, and thus one can gather an arbitrary number of measurements from a given location, time unfolds indefinitely and it is not possible to obtain measurements with the same $t$, which makes calculation of the temporal density of some phenomena difficult.

However, many events and changes occurring in human-populated environments are repetitive due to the nature of human habits.
To represent the repetitive nature of these changes, we project the time $t$ into a set of circles, where every circle is derived from the periodicity of change $T_i$ detected by FreMEn in the measured phenomenon.
This causes the time-dependent events with the same periodicity to be projected into the same areas of a circle that corresponds to the modelled periodicity, see Figure~\ref{fig:hypertime}.
The distribution of observations projected into this warped space (i.e. space with `wrapped' or `curled' time dimensions) can be then estimated using standard statistical and machine learning tools.
To predict the probabilistic distribution of a given variable for a given time $t$, we simply project $t$ to the circular space and perform the prediction there.

The projection of linear time onto a set of wrapped circles reflects the fact that human activities in the mornings of different days are more similar than human activities in the mornings and afternoons of the same day although the same day afternoon and morning is temporally closer than mornings of two different days.
Furthermore, the projection reflects the fact that a given phenomenon does not change abruptly during midnight although 23:59 and 0:01 appear to be distant.
An illustrative example of the method, which estimates $p(a|t)$ through projection into $p(a|cos(2\pi\,t/T),sin(2\pi\,t/T))$ is in Figure~\ref{fig:hypertime}.
For the sake of simplicity, this example uses only one periodicity $T$ and the spatial domain is neglected, i.e. $\mathbf{x}\in\mathbb{R}^0$.

\subsection{Method Overview}\label{sec:overview}

Let us assume that a robot gathered a set containing $l$ measurements of a given phenomenon, obtaining tuples $(a_i,\mathbf{x}_i$, $t_i)$, where $i \in \{1\ldots\l\}$, the vector $\mathbf{x}_i$ describes the location of the measurement (e.g. position of a detected person or obstacle), $t_i$ corresponds to the time of the measurement and $a_i$ represents the measurement's value, e.g. the number of detected people, likelihood of an obstacle or robot velocity in the vicinity of ($\mathbf{x}_i,t_i$).

Our method aims to find a $p(a|\mathbf{x},t)$, which would represent the conditional probability density function of the variable $a$ given the position $\mathbf{x}$ and time $t$.
\reve{The proposed method is composed of five stages: 1) initialization; 
2) spatio-temporal clustering;
3) model error estimation;
4) identification of periodicities;
5) hypertime space extension.
After initialization, our method repeatedly performs steps 2-4 as long as the model error, estimated by Eq. (\ref{eq:error}) during step 3, decreases.} 

\subsection{\reve{Step I: Initialization}}\label{sec:initialisation}
To initialize the algorithm, we first store all measurements $(a_i,\mathbf{x}_i)$ in $^h\mathbf{x}_i$, where $h$ corresponds to the number of known periodicities. Since during initialisation, the number of periodicities is unknown, we set $h$ to $0$.
In the \textit{spatio-temporal clustering} stage, we cluster the vectors $(^h\mathbf{x}_i)$, obtaining a Gaussian mixture model, which represents the spatio-temporal distribution of the given phenomenon and allows to calculate conditional probability  function $p_h(a|\mathbf{x},t)$.
In the \textit{model error estimation}, we calculate the mean $\mu_i$ of $p(a|\mathbf{x}_i,t_i)$ for all training samples.
Then we calculate the time series $^h\epsilon(t_i)$ as $^h\epsilon(t_i)= \mu_i - a_i$ and its mean squared value $E_h$, \reva{see (\ref{eq:error}) in Section~\ref{sec:error}}.
Then, during the \textit{identification of periodicities}, we \revb{use the FreMEn~\cite{fremen} method to} perform spectral analysis of $^h\epsilon(t_i)$, extract the most prominent spectral component, and store its period as $T_{h+1}$.
After that, we perform the \textit{hypertime space extension}, which extends each vector $^h\mathbf{x}_i$ by 2 dimensions representing a given periodicity of the temporal domain, i.e. 
\begin{equation}\label{eqn:extn}
^{h+1}\mathbf{x}_i \leftarrow (^h\mathbf{x}_i, \,\cos(2\pi\,t_i/T_{h+1}),\, \sin(2\pi\,t_i/T_{h+1})).
\end{equation}
Then, we increment $h$ by one and repeat the steps of \textit{spatio-temporal clustering} and \textit{model error estimation} on the now extended vectors $^h\mathbf{x}_i$, obtaining an new error $E_h$.
We compare the model error $E_h$ calculated with the error obtained in the previous iteration $E_{h-1}$ and if $E_{h} < E_{h-1}$, we proceed with \textit{identification of periodicities} and \textit{hypertime space extension}, extending the vector $^h\mathbf{x}_i$ with another two dimensions representing another potential periodicity of the modeled phenomena.
In cases where the model error starts to increase, i.e.\ if $E_{h} \geq E_{h-1}$, we store the model $p_{h-1}(a,\mathbf{x},t)$ from the previous iteration as $p(a,\mathbf{x},t)$ and terminate the method.

The resulting model allows to estimate the likelihood of each value $a$ of a given phenomena at location $\mathbf{x}$ and time $t$.
In our experiments, we show that the function $p(a,\mathbf{x},t)$ allows to predict the visibility of image features, door states, robot velocity and number of people occurrences within a given spatio-temporal volume.

   \subsection{\reve{Step II:} Spatio-Temporal Clustering}

We represent the probability density function $p(a,\mathbf{x},t)$ by a mixture of Gaussian models in the hypertime space as follows:
\begin{equation}\label{eqn:distribution}
p(a,\mathbf{x},t) = \gamma\sum_{j=1}^{n}{w_j\,u_j(a,\mathbf{x},^h\mathbf{t})},
\end{equation}
where $u_j(a,\mathbf{x},^h\mathbf{t})$ is a multivariate Gaussian function of the $j^{th}$ cluster, $w_j$ is the cluster weight, $^h\mathbf{t} = (\cos(2\pi \frac{t}{T_1}),sin(2\pi \frac{t}{T_1}),\ldots,\cos(2\pi \frac{t}{T_h}),\sin(2\pi \frac{t}{T_h}))$ is the projection of time in the hypertime space and $\gamma$ is a scaling constant, \reva{specified by (\ref{eq:scaling}) in Section~\ref{sec:error}}. 

\subsection{\reve{Step III:} Model error estimation}\label{sec:error}

Projecting the linear time $t$ onto the circular hypertime space (or its inverse) inevitably changes the scale of the calculated spatio-temporal density.
This is because several time instants $t$ can project into the same area of hypertime. 
Thus, we first need to determine the scaling factor $\gamma$ in such a way that the mean value of $a_i$ calculated from the model (\ref{eqn:distribution}) over the training set vectors $(\mathbf{x_i},t_i)$ is equal to the 
average value of $a_i$ on the training set:
\begin{equation}\label{eq:scaling}
	\reva{\gamma = \frac{\sum_{i=1}^l{a_i}}{{\sum_{i=1}^l{a_i\,\sum_{j=1}^{n}{w_j\,u_j(a_i,\mathbf{x}_i,^h\mathbf{t_i})}}}}}
\end{equation} 

After calculating the scaling factor $\gamma$, we compute an estimate of $a_i$ at each training \reve{set} point defined by location $\mathbf{x}_i$ and time $t_i$ by calculating the mean $\mu_i$: 
\begin{equation}\label{eqn:mean}
\mu_i = \int{a\,p(a,\mathbf{x}_i,t_i)}\,da
\end{equation}
Then, we calculate the error $^h\epsilon(t_i)$ as the difference between the mean and the measured values $a_i$
\begin{equation}
^h\epsilon(t_i) = \mu_i - a_i.
\end{equation}
Finally, we calculate the mean squared error of the current model as 
\begin{equation}\label{eq:error}
E_h= {\sum_{i=1}^l{^h\epsilon^2(t_i)}} = {\sum_{i=1}^l{(\mu_i - a_i)^2}}.
\end{equation}
We compare the mean squared error $E_h$ with the one calculated in the previous iteration $E_{h-1}$.
If \reva{$E_h$} is smaller than \reva{$E_{h-1}$}, we store the current model, represented by Equation (\ref{eqn:distribution}), and we perform another iteration of the method.
If \reva{$E_h > E_{h-1}$}, we terminate the algorithm.
Then, the last stored model (Eq. \ref{eqn:distribution}), can be used to predict conditional probabilistic distribution $p(a|\mathbf{x},t)$.
\reve{This is calculated in a numerical manner, i.e. we fix $\mathbf{x}$ and/or $t$ at the values we want to perform the prediction for, then we calculate $p(a_c,\mathbf{x},t)$ for values of $a_c \in \mathcal{A}$ which represent the domain of $a$, and finally we normalise the results to ensure that $\sum_{a_c \in \mathcal{A}}{p(a_c|\mathbf{x},t)} = 1$.
} 

\subsection{\reve{Steps IV and V:} Identification of periodicities and hypertime extension}\label{sec:fremenization}

To identify the periodicities in the error, we use a Fourier-transform scheme.
However, since the data collections for the experiments were performed by a system operating in real-world conditions, they were not collected in a (temporally) regular manner.
Thus, we process the time series $^h\epsilon(t_i)$ by the FreMEn method~\cite{fremen}, which, \revb{unlike traditional Discrete or Fast Fourier transforms, is suitable for finding periodicities in non-uniform and sparse data.}
In particular, we calculate the most prominent periodicity $T_h$ in the error time-series as:
\begin{equation}\label{eqn:fremen}
	T_{h+1} = arg\max_{T_k}{\sum_{i=1}^{l}{|(^h\epsilon(t_i)-^h\hat{\epsilon}) \,e^{-j\,2\,\pi\,t_i/T_k}}|},
\end{equation}
where $^h\hat{\epsilon}$ is a an average error $^h\epsilon(t_i)$. 
After establishing the $T_{h+1}$, we extend all vectors of the training set $^{h}\mathbf{x}(t_i)$ by adding another two components $(\cos(2\pi\,t_i/T_{h+1}),\, \sin(2\pi\,t_i/T_{h+1}))$, i.e. we apply Equation (\ref{eqn:extn}).

Thus, at the start of our method, each vector $^0x_i$ contains only the spatial information, i.e.\ $^0x_i = (a_i,\mathbf{x_i})$, but at the end, the vector contains $2h$ additional dimensions modelling the periodicities observed in the training data.

   \subsection{Clustering implementation}\label{sec:discussion} 

While the hypertime extension, error estimation, and periodicity estimation steps of the method are quite straightforward, deterministic and computationally inexpensive, the way to build the model of the probability distribution over the hypertime space is key to the method's predictive accurracy.
The main issue of the hypertime space is its sparsity, because the time, which is linear and one-dimensional, is projected onto a multi-dimensional \reve{hyper-torus}.
With the growing number of modelled periodicities (and thus, the number of temporal dimensions), many algorithms that we tested exhibited \reva{numerical instability}.
Thus, we dedicated a significant effort into testing various clustering methods, their initialisations and metrics \cite{kruse2007fundamentals}.

In our experiments, we utilise two clustering methods, which, in our previous experiments, provided the most satisfactory results: the HyperTime Expectation Maximisation (\textit{HyT-EM}) and HyperTime K-Means (\textit{HyT-KM}).
The \textit{HyT-EM} method is based on the Expectation Maximisation scheme implementation from the OpenCV library.
As the method requires to specify the number of clusters, we indicate the method name as \textit{HyT-EM\_k}, where $k$ is the number of clusters used during the experiments (Section~\ref{sec:experiments}).
To deal with the \reva{occasional} numerical instabilities of the OpenCV's EM implementation, the \textit{HyT-EM} method performs eigenvalue analysis of the model's covariance matrices and if necessary, it restarts the EM with different initial positions of the clusters.

The \textit{HyT-KM} method was \revc{designed to deal with the aforementioned issues, making it more suitable for clustering over the hypertime space, where the training data become sparser as the number of temporal dimensions increases}.
\textit{HyT-KM} first initialises the cluster centres using k-means based clustering. 
Inspired by~\cite{kinodynamic,roy2016swgmm,kucner2017enabling}, we use the mixtures of cosine distance for hypertime and the Euclidean metrics for other variables.
After initialisation, it calculates the covariace matrices of the clusters and then it proceeds with the Expectation Maximisation procedure, while using the cosine distance for temporal dimensions.
\revb{In contrast to} \textit{HyT-EM}, \textit{HyT-KM} does not require to specify the number of clusters in advance.
Instead, the algorithm tries to analyse the temporal structure of the hypertime space prior to the \textit{hypertime expansion} step.
In particular, it starts with $n=1$ clusters.
It builds models with $n$ and $n+1$ clusters and calculates the sum of amplitutes $T_{\Sigma}(n)$ and $T_{\Sigma}(n+1)$ of the frequency spectrum of the error $^h\epsilon(t_i)$
\begin{equation}
T_{\Sigma}(n) = \sum_{k=1}^{K}{\sum_{i=1}^{l}{|(^h\epsilon(t_i)-^h\hat{\epsilon}) \,e^{-j\,2\,\pi\,t_i/T_k}}|}.
\end{equation}
If $T_{\Sigma}(n) > T_{\Sigma}(n+1)$, the model with $n+1$ clusters is stored and the number of clusters $n$ is incremented by 1.
If $T_{\Sigma}(n) \leq T_{\Sigma}(n+1)$, the method simply proceeds with the \textit{hypertime expansion} step.

   \section{Experiments}\label{sec:experiments}

The purpose of the experimental evaluation is to assess the predictive capability of the proposed method and its utility for different robotic tasks.
The performance of the method is evaluated in four different scenarios, which require predictions of variables of different dimensionalities.
The data for these experiments were collected by robotic sensors in real world conditions over periods of several weeks.
These scenarios relate to a real deployment of a mobile robot in human populated environment, see Section~\ref{sec:overview} and \cite{hawes2016strands}, where we discussed how a long-term operating robot will benefit from the predictive capabilities of models that explicitly represent temporal behaviour of environment states with different dimensions.
To evaluate the efficiency of our method, we compare five different temporal models: \textit{Mean}, which predicts a value as an average of its past measurements, \textit{Hist\_n}, which divides each day into $n$ intervals and predicts the given variable as an average in a relevant time of a day, \textit{FreMEn\_m}, which extracts $m$ periodic components from the variable's history and uses these periodicities for prediction, \textit{HyT-EM\_k}, which uses the expectation-maximisation of $k$-component Gaussian Mixture Model over the hypertime space, and finally \textit{HyT-KN}, as described in Section~\ref{sec:discussion}.
The experimental evaluation is performed by an automated system~\cite{evaluation}, which first optimises each method's parameters (number of intervals $n$, number of periodicities $m$, and number of clusters $k$) and then runs a series of pairwise t-tests to determine which methods perform  statistically significantly better than other ones \reva{in terms of their prediction error}.
The results of the statistical evaluations are shown on the right sides of the Figures~\ref{fig:binary}-\ref{fig:pedestrians}, where an arrow from A to B indicates that method A achieved statistically significantly \reva{less erroneous predictions} than method B.
To enable the reproducibility of the results, the evaluation system, source codes and datasets are available online~\cite{fremen-www}.

\subsection{Door state}

The first scenario concerns a single binary variable, which corresponds to the state of a university office door.
The door was continuously observed by an RGB-D camera for 10 weeks to obtain the training set, and for another 10 weeks to obtain 10 testing sets, each one week long.
Since the RGB-D data processing was rather simple, the data contains noise, because people moving through the door caused the system to indicate incorrectly that the door was closed.

To compare the efficiency of the predictions, we calculated the mean squared error $\epsilon$ of the various temporal models' predictions $p(t)$ to the ground truth $s(t)$ as $\epsilon={\sum_T{(p(t)-s(t))^2}/|s(t)|}$.
\begin{figure}[!ht]
   \begin{center}
      \hfill
      \includegraphics[height=3.0cm]{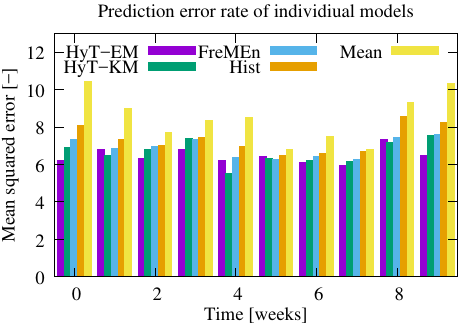}
      \hfill
      \includegraphics[height=3.0cm]{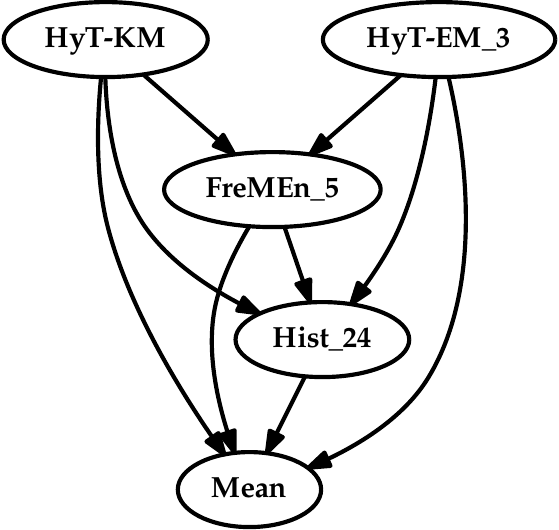}
      \hfill
      \caption{Door state prediction error. The left figure shows the MSE for the training (week 0) and testing (weeks 1-9) datasets. An arrow from model A to model B in the right figure indicates that A's prediction error is statistically significantly lower than prediction error of model B.\label{fig:binary}}
   \end{center}
	\figurespace
\end{figure}
The results shown in Fig.~\ref{fig:binary} indicate that both hypertime-based models outperformed the other ones, including FreMEn~\cite{fremen}.
\revd{Similarly to the results of~\cite{fremen}, both methods indicated that that the best predictions are achieved by modeling three most prominent periodicies. 
In this experiment, these corresponded to one day, four hours and one week.} 
%As shown in~\cite{17}, the accuracy of predictions provided by our method allows for efficient anomaly detection.

\revd{The reason behind the superiority of the hypertime-based methods is that FreMen approximates the training data by only one (sinus) function per observed periodicity. 
This causes difficulties when modeling states, which are influenced by several processes with the same periodicity or when representing short duration, regular events.
The hypertime-based methods combine the advantages of mixture-based representations, which can approximate arbitrarily-shaped, multimodal functions, with the ability of FreMEn to represent their natural periodicities.
For an illustrative example of the advantages of GMM and FreMEn models, leading to their combination in this work, please see our earlier work~\cite{fremen_search}.
}

\subsection{Topological localisation}

In this scenario a robot has to determine its location in an open-plan university office based on the current image from its onboard camera and a set of pre-learned appearance models of several locations.
Since the appearance of these locations changes over time, it is beneficial to utilise appearance models that explicitly represent the appearance variations~\cite{prediction,fremen,rosen,churchill}.
This experiment compares the impact of different temporal models, which predict the visibility of environmental features at these locations, on the robustness of robot localisation. 
To gather data about the changes in feature visibility, a SCITOS-G5 robot visited eight different locations of the university office every 10 minutes for one week, collecting a training dataset with more than 8000 images.
After one week, the robot visited the same locations every 10 minutes for one day, collecting 1152 time-stamped images used for testing. 
% 
%\begin{figure}[!ht]
%   \begin{center}
%      \includegraphics[width=0.99\columnwidth]{fig/brayford.jpg}
%%%      \caption{Example images of the indoor training dataset. Shows the appearance of six monitored locations on November 2013.\label{pic:places}}
%   \end{center}
%\end{figure}
%
The training set images were then processed by the BRIEF method~\cite{brief}, which shows good robustness to appearance changes~\cite{krajnik2016griefras}.
The extracted features belonging to the same locations were matched and we obtained their visibility over time, which was then processed by the temporal models evaluated.
Thus, we obtained a dynamic appearance-based model of each location that can predict which features are likely to be visible at a particular time.

\begin{figure}[!ht]
   \begin{center}
   \hfill
      \includegraphics[height=3.0cm]{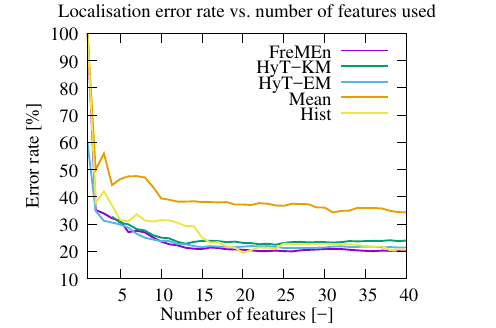}
      \hfill
      \includegraphics[height=2.95cm]{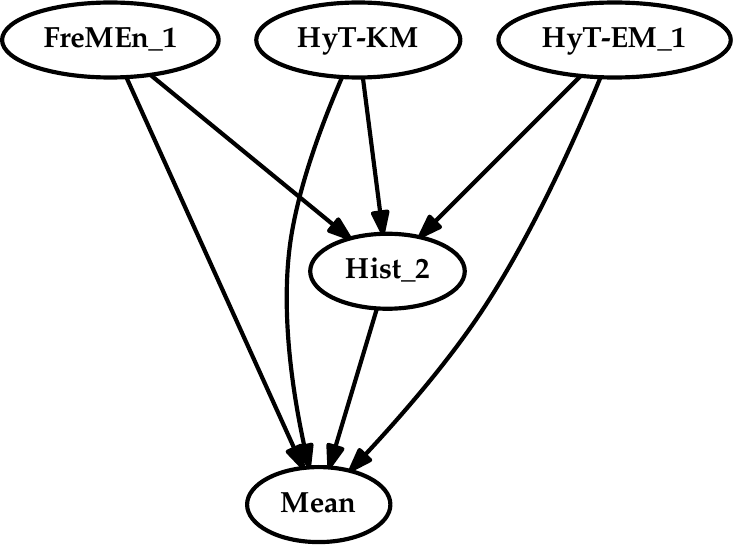}
      \hfill
      \caption{Temporal model performance for feature-based topological localisation. The left figure shows the dependence of localisation error rate on the number of features predicted by a given temporal model.
An arrow from A to B in the right indicates that A's localisation error rate is statistically significantly lower than localisation error rate of model B.\label{fig:features}}
   \end{center}
	\figurespace
\end{figure}

During testing, the robot uses these models to calculate the likelihood of the features' visibility at each of the locations at the time it captured an image by its onboard camera (or extracted a time-stamped image from the testing set).
In particular, it selects the $n$ most likely-to-be-visible features at each location and time, matches these features to the features extracted from its onboard camera (or testing set) image, and determines the model with the most matches as its current location.
The localisation error is calculated as the ratio of cases when the robot incorrectly estimated its location to the total number of images in the testing set. 
The dependence of the average localisation error on the particular temporal model and number of features $n$ used for localisation is shown in Figure~\ref{fig:features}.
The results indicate that the localisation robustness of the methods that take into account the rhythmic nature of the appearance changes outperform the \textit{Mean} method, which relies on the most stable image features.
Moreover, the methods that model these cyclic changes in a continuous manner perform better than the \textit{Hist} method which models different times of the day in separate, as shown in Figure~\ref{fig:features}. 

\subsection{Velocity prediction}

This scenario concerns the ability of our representation to predict the velocity of a robot while navigating through a given area, which depends on how cautiously it has to move due to the presence of humans.
Thus, this experiment is concerned with the ability of our method to predict a one-dimensional continuous variable (robot velocity) for a given time and location. 

The velocities and times of navigation for our evaluation were obtained from a database obtained with a SCITOS-G5 mobile platform, which gathered data in an open plan research office for more than 10 weeks.
\revb{Since the robot was used for other purposes, the data were not gathered in a regular way, but contain long hiatuses, see~\cite{7}.}
Typically, the average velocity of the robot did not show much variation, but in cases it had to navigate close to workspaces and through doors, the velocity varied significantly.
To evaluate the ability of our approach to predict the robot velocity, we split the dataset into an 8-weeks long training set and two testing sets of 1-week duration.
\begin{figure}[!ht]
   \begin{center}
   \hfill
      \includegraphics[height=2.9cm]{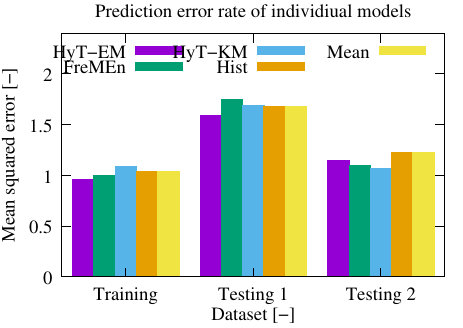}
      \hfill
      \includegraphics[height=2.5cm]{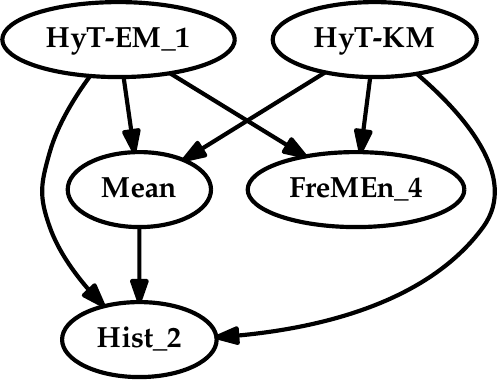}
	   \caption{Navigation velocity \reva{reconstruction and} prediction errors. The left figure shows the MSE for the training and testing sets.
	   An arrow from A to B in the right indicates that A's prediction error is statistically significantly lower than velocity prediction error of model B.\label{fig:navigation}}
   \end{center}
	\figurespace
\end{figure}
As in the case of door state prediction, we calculated the mean square error of the predictions provided by our models, and compared them to find out which of the methods provide the most accurate predictions.
\reva{While the HyT-KM reconstruction error (i.e. error calculated on the training data) is higher, its prediction error is lower compared to \textit{Mean}, \textit{FreMEn}  and \textit{Hist} methods, see Figure~\ref{fig:navigation}}.
\revc{Since the velocity of the robot is always between 0 and 1, we applied FreMEn directly to the velocity values $v\in<0,1>$ and considered the FreMEn prediction for a particular time a velocity. The same scheme was used to predict the robot velocity in~\cite{hawes2016strands}.}

\subsection{Human presence}

Finally, we validated the proposed approach on 2-dimensional data indicating the positions of people in several corridors of the Isaac Newton Building at the University of Lincoln.
Data collection was performed by a mobile robot equipped with a Velodyne 3d laser rangefinder, which was placed at a T-shaped junction so that its laser range-finder was able to scan the three connecting corridors simultaneously.
To detect and localize people in the 3d point clouds provided by the scanner, we used an efficient and reliable person detection method~\cite{yan2017online}.
Since we needed to recharge the robot occasionally, we did not collect data on a 24/7 basis and recharged the robot batteries during nights, when the building is vacant and there are no people in the corridors.
\revb{Similarly to the previous case, the robot was often used for other purposes, resulting in large gaps in the data collected.}
Thus, our dataset spanned from early mornings to late evening over several weekdays.
Each day contains approximately $28000$ entries, which correspond to hundreds of walks by people through the monitored corridor. 
\begin{figure}[!ht]
   \begin{center}
   \hfill
      \includegraphics[height=3.0cm]{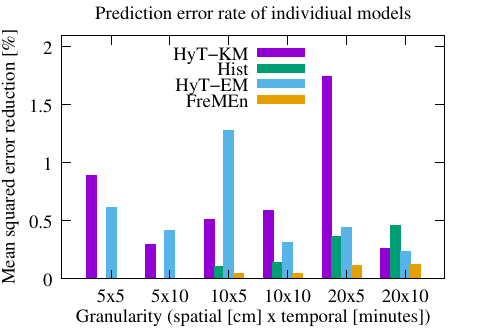}
      \hfill
      \includegraphics[height=2.8cm]{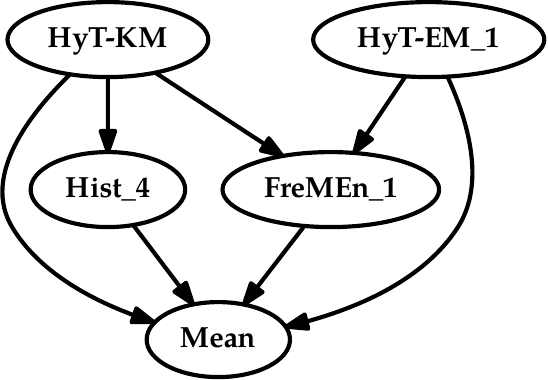}
      \hfill
	   \caption{Human presence prediction results. The left figure shows how the mean squared error reduced for a particular model \reva{granularity} compared to the \textit{Mean} model.
An arrow from A to B in the right indicates that A's prediction error is statistically significantly lower than velocity prediction error of model B.\label{fig:pedestrians}}
   \end{center}
	\figurespace
\end{figure}
To quantitatively evaluate the model quality, we again split the gathered data into training and test sets, and learn the model from the training set only.
Then, we partition the timeline of the test data into a spatio-temporal 3d grid.
For each cell $g$, we count the number of detections $d_g$ that occurred and compare this value with the value $p_g$ predicted by a given spatio-temporal model.
To better visualise the methods' prediction improvements, we show the reduction of the mean square error compared to the $Mean$ model in Figure~\ref{fig:pedestrians}.
To make a comparison with other models, we apply the FreMEn method on each of the grid cells independently and then predict the most likely number of events at a given time in a particular cell.
Since the error is dependent on the partitioning used, we tested the method for grids of various cell sizes ranging from 5 to 20 cm and 5 to 30 minutes.
\revc{Since the number of people passing through a cell follows a Poisson distribution, we employed FreMEn modification proposed in~\cite{15}, which allows to process non-binary values.
However, the number of passing people exceeds one only in rare occations, and one would obtain almost the same results by using the original FreMEn~\cite{fremen}.}

To demonstrate the model's ability to estimate the spatio-temporal distribution over time, we let it predict the most likely occurrence of people for different times and composed a video~\cite{video} showing how the predicted distributions of the people depend on time. 

%
%\begin{figure}[!t]
%   \begin{center}
%      \includegraphics[width=0.45\columnwidth]{fig/corridor0}
%	\hfill
%      \includegraphics[width=0.45\columnwidth]{fig/corridor1}\\\vspace{4mm}
%      \includegraphics[width=0.45\columnwidth]{fig/corridor2}
%	\hfill
%      \includegraphics[width=0.45\columnwidth]{fig/corridor3}
%      \caption{Spatio-temporal model of people presence in a corridor of the University of Lincoln, UK at various times of a day. See a video at~\cite{video}\label{fig:spatemp}.}
%   \end{center}
%\end{figure}
%

   \section{Conclusion}

We presented a novel approach for spatio-temporal modeling for robots that are required to operate for long periods of time in changing environments.
The method models the time domain in a multi-dimensional hyperspace, where each pair of dimensions represents one periodicity observed in the data.
This multi-dimensional, warped time model is used to extend the state space representing a given phenomenon.
By projecting the robot's observations into this space-hypertime and clustering them, we create a continuous, spatio-temporal model (distribution) of the phenomenon observed by the robot.
Knowledge of the spatio-temporal distribution is then used to predict the probabilistic distribution of a given phenomenon at a given time and location.

Using data collected by a mobile robot over several weeks, we show that the method can represent the spatio-temporal dynamics of binary and continuous variables, and use the representation to make predictions of the future environment states, resulting in significantly better performance than the previous state of the art.

\revd{The main contribution of the article does not rest in the clustering methods, but in the alternative representation of the time domain, which, unlike the linear time, respects the repetitiveness of events and changes imposed by naturally-occurring cycles.}
Moreover, we outline a formulation of the problem, which allows to apply the aforementioned method to most environmental models used in the robotic domain.
\revc{We believe that this human-oriented circular representation of time will allow service robots to synchronise their activities with the habits of people they are supposed to serve.} 

One of the major problems we have encountered is the instability of clustering methods on the multi-dimensional hypertime space.
Thus, in the future, we will evaluate the performance of different clustering algorithms on the proposed representation \reve{considering models capable of treating spatial and temporal domains in separate or employing only a subset of the detected periodicities~\cite{huang}.}
Furthermore, the observed distribution is heavily influenced by the way in which the robot samples the environment, i.e.\ where and when it takes the measurements.
Thus, we will study spatio-temporal exploration methods, which will allow a mobile robot to automatically select a location and time to obtain data useful to refine and improve the spatio-temporal model. 
\revc{Moreover, we will extend the spatio-temporal modeling approach towards its ability to take into account sensor noise in way similar to standard Bayesian update applied to states with fixed probabilistic values.}

To allow use of the method by other researchers, we provide its baseline open source code and datasets at \reve{\url{chronorobotics.tk}} and \url{fremen.uk}.

\bibliographystyle{IEEEtran}
\bibliography{main}

\end{document}